\documentclass[]{llncs}
\usepackage[T1]{fontenc}

\usepackage{times}
\usepackage{helvet}
\usepackage{courier}
\usepackage{graphicx}
\usepackage{amsmath}
\usepackage{amssymb}
\usepackage{hyperref}
\usepackage{amsmath}
\usepackage{booktabs} 
\usepackage{siunitx} 
\usepackage{tabularx} 
\usepackage{array} 
\usepackage{todonotes}
\usepackage{makecell} 
\usepackage{caption}

\newcolumntype{Y}{>{\centering\arraybackslash}X}
\newcolumntype{M}[1]{>{\centering\arraybackslash}m{#1}}

\begin{document}
\captionsetup[table]{skip=10pt}

%
\title{Advancing EEG-Based Gaze Prediction Using Depthwise Separable Convolution and Enhanced Pre-Processing}
\author{Matthew L Key\textsuperscript{*} \and Tural Mehtiyev\textsuperscript{*} \and Xiaodong Qu}

\institute{Department of Computer Science\\
The George Washington University\\
Washington DC, USA}

\maketitle
\renewcommand{\thefootnote}{\fnsymbol{footnote}}
\footnotetext[1]{The first authors contributed equally to this work.}
\footnotetext[2]{Full source code is available at \href{https://github.com/GWU-CS/EEG-DCViT}{https://github.com/GWU-CS/EEG-DCViT}}

\begin{abstract}
\begin{quote}
In the field of EEG-based gaze prediction, the application of deep learning to interpret complex neural data poses significant challenges. This study evaluates the effectiveness of pre-processing techniques and the effect of additional depthwise separable convolution on EEG vision transformers (ViTs) in a pretrained model architecture. We introduce a novel method, the EEG Deeper Clustered Vision Transformer (EEG-DCViT), which combines depthwise separable convolutional neural networks (CNNs) with vision transformers, enriched by a pre-processing strategy involving data clustering. The new approach demonstrates superior performance, establishing a new benchmark with a Root Mean Square Error (RMSE) of 51.6 mm. This achievement underscores the impact of pre-processing and model refinement in enhancing EEG-based applications.
\newline

\textbf{Keywords:} EEG, Gaze Prediction, Machine Learning, Vision Transformer, EEGEyeNet, Depthwise Separable Convolution.

\end{quote}
\end{abstract}

\subsection{1   Introduction}

Electroencephalogram (EEG) data, with its multidimensional architecture, captures an abundance of details regarding brain functions, providing various perspectives on numerous neurological events \cite{murungi2023trends,qu2020using,teplan2002fundamentalseeg}. 
Despite the widespread use of machine learning regression models for EEG data, their complexity and expensive data collection process often hinder these models from effectively understanding the data's complex structures \cite{craik2019deeplearningeeg,qu2020multi}. The EEGEyeNet dataset, with its extensive collection of EEG and eye tracking (ET) data, emerges as a significant asset in this field, enabling in-depth gaze behavior study and laying the groundwork for benchmarking gaze prediction approaches \cite{kastrati2021eegeyenet}. Leveraging the EEGEyeNet dataset, the hybrid vision transformer (ViT) has showcased its potential in gaze prediction, challenging conventional convolution-based approaches \cite{midterm-eeg-vit}. As a contribution to the field, our study delves into how alterations in EEGViT design with additional depthwise separable convolution, combined with pre-processing techniques, can amplify the accuracy in predicting absolute eye position. Following our findings, we propose a new model which obtains better than state of the art performance on EEGEyeNet abosolute eye position.

\subsection{1.1   Research Questions}

To further elucidate our direction within this evolving landscape, we formulate two pivotal research questions (RQs):

RQ 1: In what ways does incorporating depthwise separable convolution into EEG-based gaze prediction models influence their predictive accuracy?

RQ 2: What impact do advancements in pre-processing techniques have on the accuracy of EEG-based gaze prediction models?

\subsection{2   Related Work}

Gaze prediction, with its extensive applications in human behavior analysis, advertising, and human-computer interactions, has garnered significant attention. The traditional reliance on Convolutional Neural Networks (CNNs) for this task has been reconsidered due to their limitations in capturing complex EEG patterns \cite{chaaraoui2012review,yi2022attention}\cite{majaranta2014eyetracking}\cite{okada2023advertisement}.  The introduction of the EEGViT model, incorporating Transformer blocks, represents a significant shift, offering a promising alternative to conventional convolutional approaches \cite{midterm-eeg-vit}.

The integration of ViTs with EEG-based gaze prediction marks a notable advancement, utilizing deep learning to navigate the intricacies of brain data interpretation. The effectiveness of both pure and hybrid transformer models in gaze estimation has been showcased, illustrating their capability in extracting detailed spatial features \cite{cheng2021gaze}. Such models demonstrate the versatility of transformers, adapting well across different data modalities and enhancing the accuracy of human gaze prediction.

Transformers have also been applied beyond gaze prediction, notably in EEG signal analysis for tasks like epileptic seizure prediction. This broadens the scope of transformer applications from their origins in NLP to encompass the analysis of temporal and spatial EEG signal features, highlighting their adaptability and potential in handling complex EEG data \cite{godoy2022eeg}.

Furthermore, exploring the synergy between CNNs and transformers has opened new avenues for EEG data processing. This combined approach leverages CNNs for local feature extraction and transformers for global dependency modeling, as demonstrated in Transformer-guided CNNs for seizure prediction. Such innovations underline the potential of integrating CNN and transformer architectures to achieve higher accuracy and better generalization in EEG-based applications, including gaze prediction \cite{godoy2022eeg} \cite{cheng2021gaze}.

This evolving landscape underscores the promise of combining CNNs and transformers in EEG data analysis, guiding our research towards optimizing such integrations. By harnessing the strengths of both architectures, we aim to set new standards in EEG-based gaze prediction and neural data interpretation, contributing to the field's advancement.

\begin{figure}
    \centering
    \includegraphics[width=.75\linewidth]{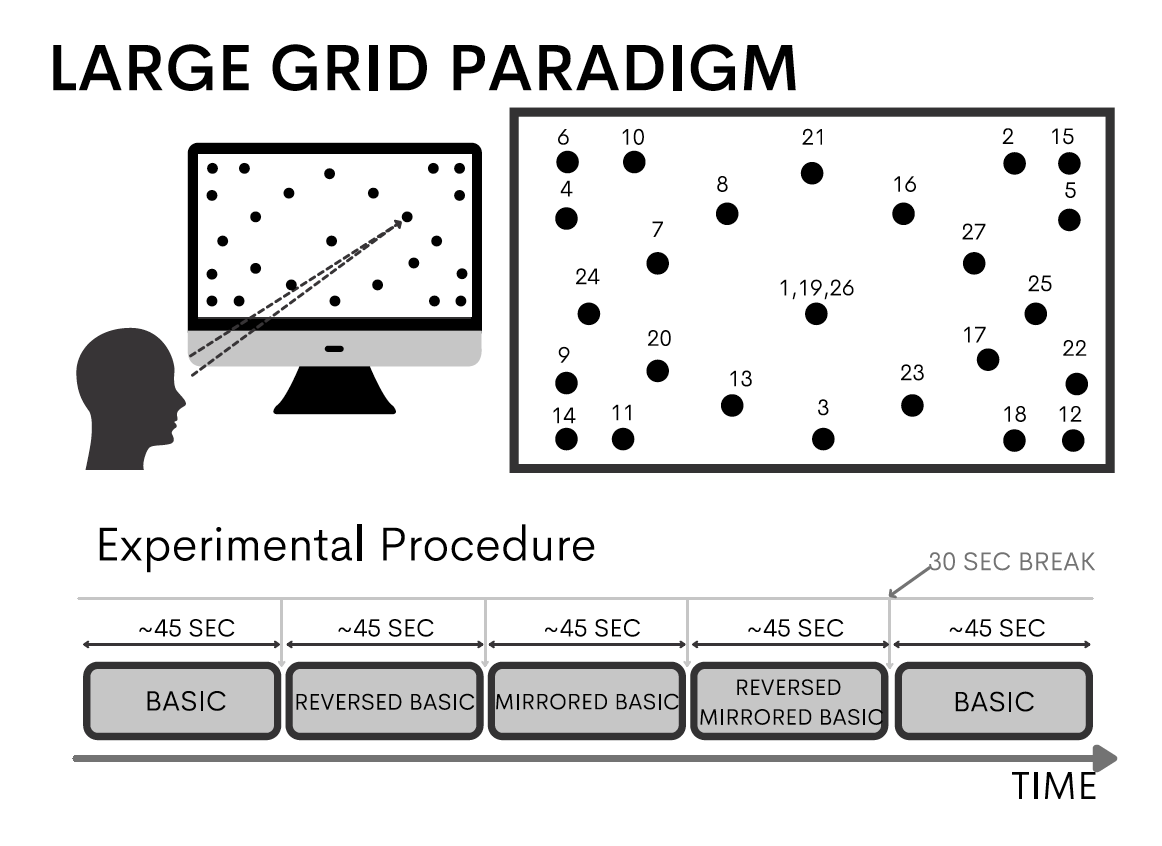}
    \caption{\textbf{Large Grid Experimental Setup:} This image illustrates the schematic view of the experimental setup and the stimuli placement on the screen. It gives a visual representation of how participants interacted with the stimuli during the eye-tracking events \cite{kastrati2021eegeyenet}.}
    \label{fig:Large Grid Paradigm}
\end{figure}

\subsection{3   Methods}

Our research extends the work presented in \cite{midterm-eeg-vit}, focusing on the utilization of pre-processing and depthwise-separable convolution techniques in EEG-based gaze prediction methodologies.

\paragraph{Data Pre-Processing:}

\begin{figure}
    \centering
    \includegraphics[width=.75\linewidth]{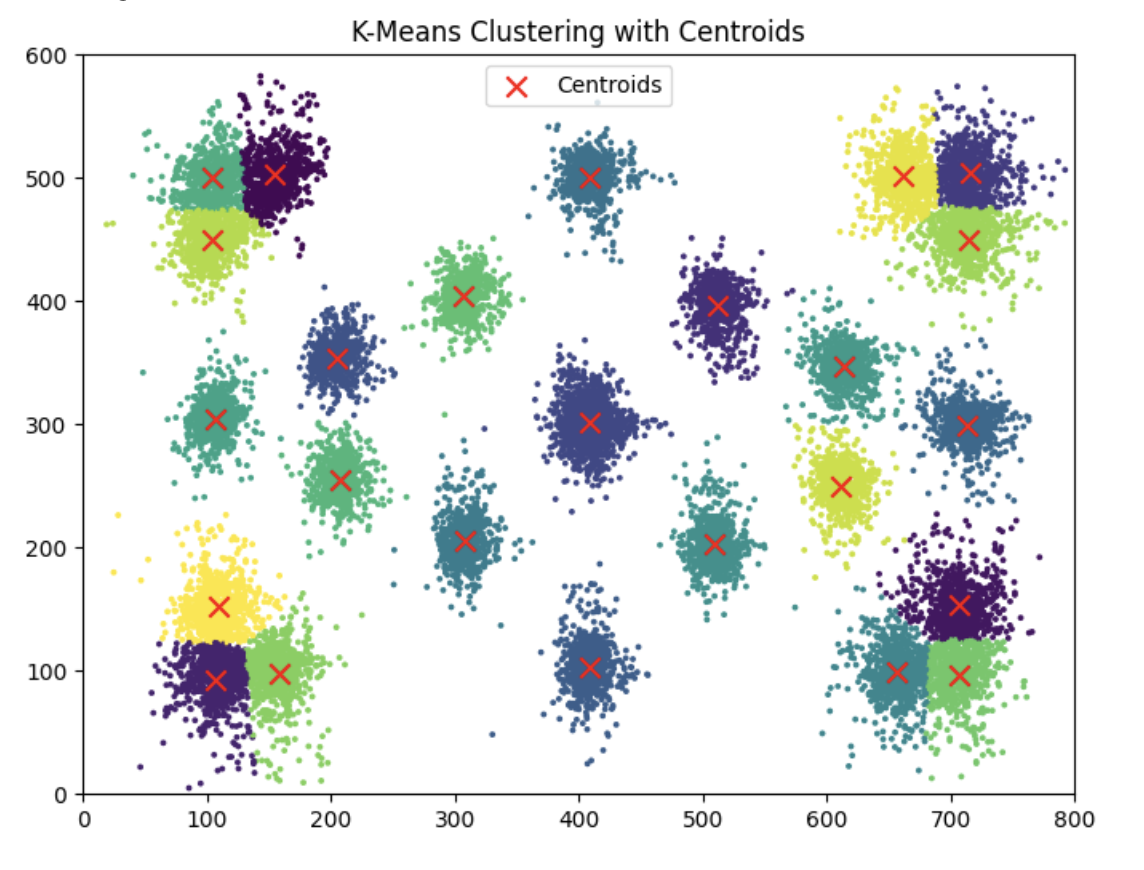}
    \caption{Clustering illustrates the discrepancy between labeled positions and actual target positions.}
    \label{fig:clustering}
\end{figure}

Pre-processing techniques have become crucial in enhancing the performance of pre-trained vision transformer models, as noted in studies by Chen et al. (2021) \cite{chen2021pretrained} and Li et al. (2021) \cite{li2021efficient}. In our analysis of the EEGEyeNet dataset, we noted the presence of significant noise. During the original data collection, the EEGEyeNet procedures required participants to focus on specific target positions. Kastrati et al. (2021) \cite{kastrati2021eegeyenet} reported that, with the computer monitor used in the experiment, 1 pixel equates to 0.5 mm. However, we identified x and y label positions in the dataset that are as much as 100 pixels (or 50 mm) away from any known target position (Figure \ref{fig:clustering}). This significant discrepancy led us to hypothesize that participants were indeed looking at the target positions, suggesting a potential issue with the eye-tracking system. This inaccuracy leads to inherent biases in the label positions which cannot be learned during model training. These errors could be the result of the system's malfunction or improper calibration.

Another potential source of error might stem from the disparity in the granularity of the data collected. The EEG data were captured at a frequency of 500 Hz, equivalent to 500 times per second. In contrast, the eye-tracking data were recorded at a much lower frequency, once per second \cite{kastrati2021eegeyenet}. Therefore, if a participant's gaze was in transit towards a target point when captured, the recorded eye position might not accurately represent the entire second during which the brainwave data were collected. Unfortunately, with the available data, it is impossible to determine the exact position of the participant's eyes throughout each sample.

\begin{figure}
    \centering
    \includegraphics[width=.75\linewidth]{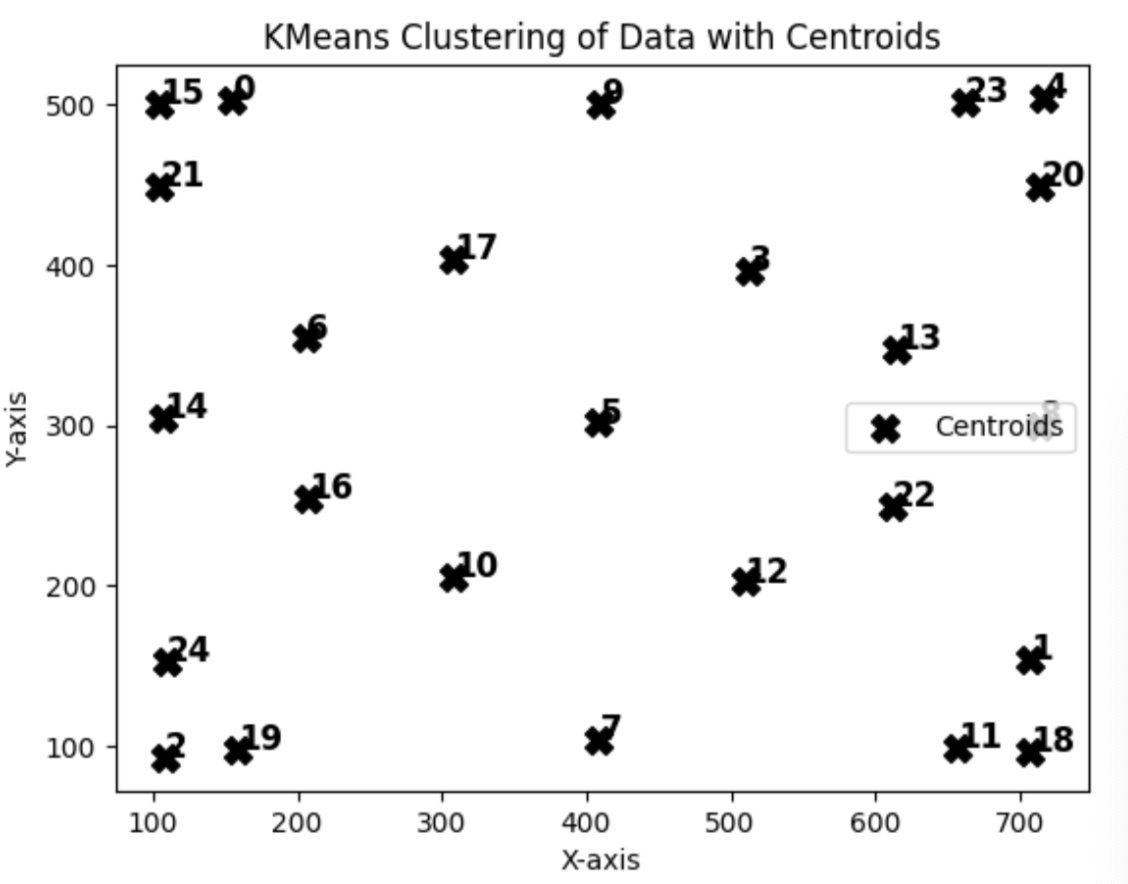}
    \caption{The centroids used to correct training data labels.
}
    \label{fig:The centroids}
\end{figure}

To address the discrepancy in eye-tracking location, we employed K-means clustering to reconcile the differences between the labeled position and the actual target position. By updating the true label position with the centroids, as illustrated in Figure \ref{fig:The centroids}, we aligned it with the cluster center position, thereby enhancing the accuracy of our dataset.

\paragraph{Depthwise-separable convolutional neural networks (DS-CNNs):}

 Early studies \cite{krizhevsky2012imagenet}  highlighted that initial layers of CNNs are adept at detecting edges or specific colors in natural images. In recent years, research aiming to gain a deeper understanding of how Convolutional Neural Networks (CNNs) operate has largely shifted towards analyzing the features learned by convolutional layers rather than the weights themselves \cite{zeiler2013visualizing} \cite{yosinski2015understanding}.
  While examining the learned features of convolutional layers is a logical approach, the interpretation of the filter weights in the deeper layers of CNNs remains a challenge. Meanwhile, Depthwise-Separable Convolutional Neural Networks (DS-CNNs) have been rising in prominence within the field of computer vision and demonstrated state-of-the-art accuracy while requiring significantly fewer parameters and computational operations than traditional CNNs, owing to the reduced computational demands of DS-CNNs \cite{howard2017mobilenets}.

The application of depthwise separable convolution in EEG data analysis shows its potential in enhancing model performance through efficient feature extraction from multichannel EEG signals. The high accuracy rates achieved in emotion recognition tasks using publicly available EEG datasets, as cited in the works by Li et al. \cite{li2022multidimensional} and further supported by studies \cite{huang2023cdba}, \cite{wang2023decoding}, underscore its effectiveness in reducing computational load while maintaining or improving performance score.

Building on these findings, we extend the application of depthwise separable convolution to the EEGEyeNet dataset. EEGEyeNet, being a comprehensive dataset for gaze estimation and other EEG-based analyses, could benefit significantly from the effective feature extraction capabilities of depthwise separable convolution. This approach may enhance the accuracy, especially in tasks requiring the analysis of spatial EEG signal characteristics. The potential for improved performance in EEG-based predictive modeling with reduced computational demands makes depthwise separable convolution a promising technique for exploration in this dataset.

 We apply depthwise separable convolution by expanding the the previous work \cite{midterm-eeg-vit} where the authors developed a hybrid vision transformer architecture named EEGViT, specifically tailored for EEG analysis. This model integrates a traditional two-step convolution operation during the patch embedding process. The first step involves a convolutional layer employing a 1×T kernel to capture temporal events across channels, acting as band-pass filters for EEG signals. Following this, the second step involves a depthwise convolutional layer with a C×1 kernel, designed to filter inputs across multiple channels at the same point in time. The model segments input images into C×T patches, which undergo a row-by-row linear projection, transforming each column vector into a scalar feature.

\begin{figure}
    \centering
    \includegraphics[width=1\linewidth]{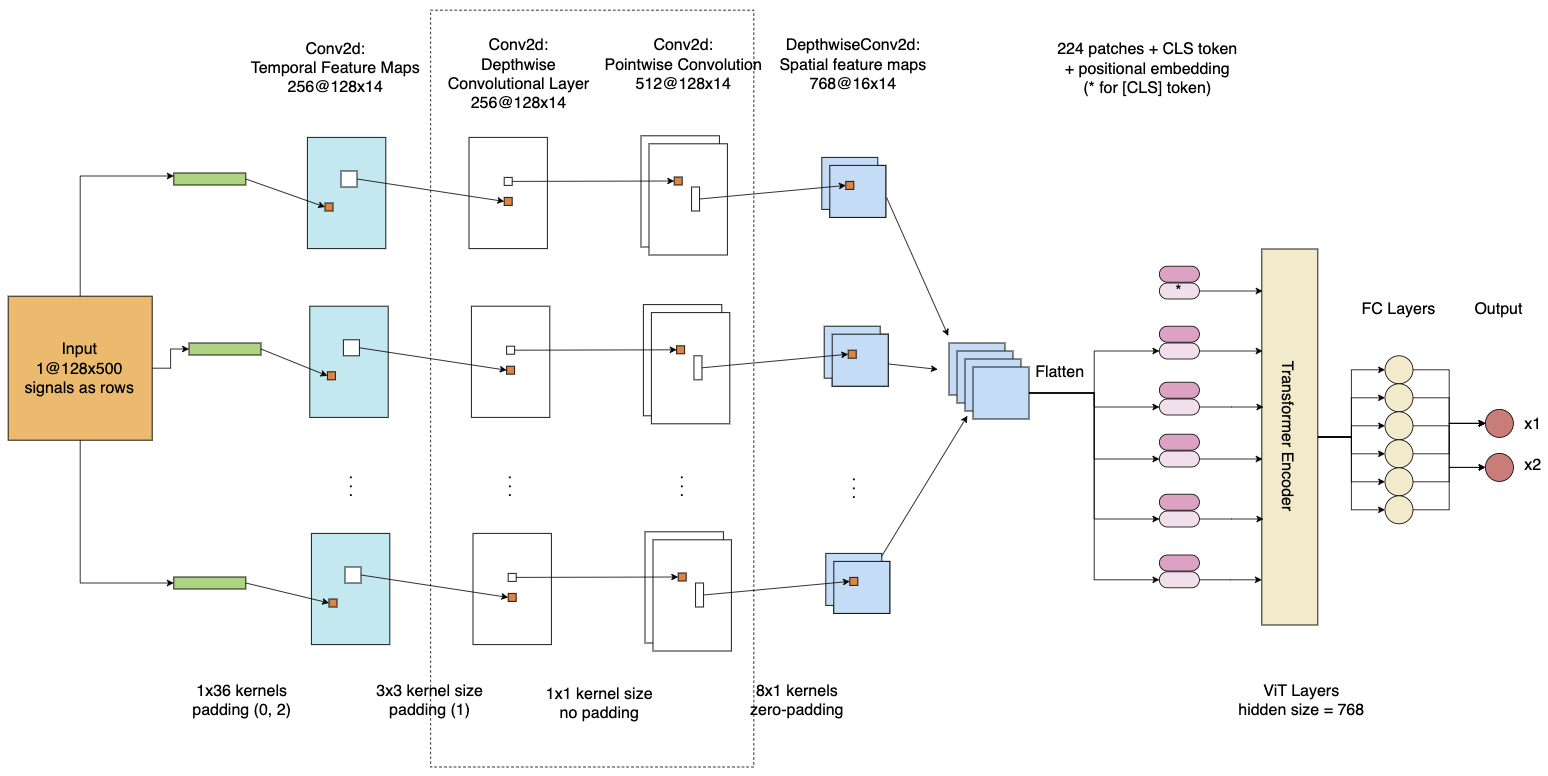}
    \caption{\textbf{EEG Vision Transformer with Depthwise Separable Convolution} A specialized ViT structure tailored for raw EEG signal input. This architecture utilizes a quad-step convolution process to produce patch embeddings. The dotted outline highlights the depthwise separable convolution. After this initial step, positional embeddings are integrated and the combined sequence is subsequently passed through the ViT layers \cite{midterm-eeg-vit}. The design of the positional embedding and ViT layer is adapted from \cite{dosovitskiy2021image}.
}
    \label{fig:eeg-vit-architecture}
\end{figure}

Building on previous study, we introduce an additional depthwise separable convolution layer in our approach. This layer incorporates both depthwise and pointwise convolutions. Following this enhancement, as shown in figure \ref{fig:eeg-vit-architecture}, our systematic approach for EEG data classification begins with a 2D convolution layer employing 256 filters of size (1, 36), featuring a stride of (1, 36) and padding of (0, 2). This layer is tasked with extracting temporal features from EEG signals. Subsequently, the depthwise separable convolution layer, comprising 256 filters for the depthwise part and 512 filters for the pointwise part, processes spatial information across channels. The architecture further integrates a ViT, modified with a custom depthwise convolution layer using 512 filters of size (8, 1). The process concludes with a classifier that includes a linear layer, a dropout layer, and a final linear layer, responsible for outputting logits that indicate class probabilities in a binary classification task. The incremental addition of the depthwise separable convolution layers in on the previous approach has proven to be effective in generating enhanced spatial features. These improved features effectively contribute to the model's ability to refine its performance and improve its accuracy.

\begin{table}[h]
\centering
\begin{tabularx}{0.5\textwidth}{|M{0.2\textwidth}|Y|}
 \hline
 \textbf{Method} & \textbf{Description} \\
 \hline
 Method 1 & EEGViT Trained with DS-CNNs \\
 \hline
 Method 2 & EEGViT Trained with Clustered Data \\
 \hline
 Method 3 (EEG-DCViT) & EEGViT Trained with Clustered and DS-CNNs \\
 \hline
\end{tabularx}
\caption{Descriptions of the methods used in the study.}
\label{table:methods}
\end{table}

\paragraph{Evaluation Metrics:}
To maintain consistency and ensure comparability with prior work, all methods, whether applied individually or in combination, will be gauged using the root mean squared error (RMSE).

\paragraph{Early Stopping:}
We employ a type of early-stopping during training to improve model performance. The SOTA EEG-ViT model was trained on a static number of 15 epochs \cite{midterm-eeg-vit}. However, the authors did not take advantage of the validation set to detect when the model was overfitting to the training data. During training, our algorithm run for 15 epochs and then output the trained model based on the epoch that has the best validation score. This will protect against overfitting and encourage higher overall accuracy.

As outlined in Table \ref{table:methods}, our study employs several methods to address the problem at hand. Each method has been tailored to optimize performance based on the specific characteristics of the dataset and the goals of the analysis.

\paragraph{Method 1: EEGViT Trained with DS-CNNs: } 
This approach leverages a pre-trained EEGViT model, further refined using depthwise-separable convolutional neural networks as an additional layer. Known for their superior spatial feature extraction capabilities, DS-CNNs enable the model to effectively identify and process complex patterns in EEG channels. This method addresses Research Question 1 by demonstrating the impact of depthwise separable convolutions techniques on the accuracy score in EEG data.
    
\paragraph{Method 2: EEGViT Trained with Clustered Data: }
By clustering the data prior to training, we can ensure that the model is exposed to the most representative and diverse examples. This pre-processing step helps in improving the generalization capability of the EEGViT model by focusing on the underlying distribution of the dataset.  This method addresses Research Question 2 by exploring the impact of data processing step on the model performance in EEG data.
    
\paragraph{Method 3 (EEG-DCViT): EEGViT Trained with Clustered and DS-CNNs:}
This method, EEG Deeper Clustered Vision Transformer (EEG-DCViT), integrates the techniques of data clustering with depthwise separable convolutional neural networks (DS-CNNs) to harness the advantages of both approaches. By clustering the EEG data, the model can focus on learning from more homogeneous subsets, which improves its efficiency in recognizing underlying patterns. When combined with the DS-CNNs, known for their enhanced feature extraction with fewer parameters and computational efficiency, this strategy significantly boosts the model's capacity to identify intricate and subtle patterns within the EEG channels. This dual approach integrates the findings from both research questions to enhance the training phase, laying a robust foundation for the model. This integration aims to boost the accuracy and improve the generalization capabilities of EEG data analysis.

\subsection{4   Dataset}
The EEGEyeNet dataset comprises data from 27 participants with a total of 21,464 samples \cite{kastrati2021eegeyenet}. The primary focus is on the "Absolute Position" task where the objective is to ascertain the exact gaze position in terms of XY-coordinates on the screen. Each sample corresponds to a one-second duration where a participant engages in a single fixation on the Large Grid paradigm (Figure \ref{fig:Large Grid Paradigm}). The performance is assessed by measuring the Euclidean distance between the actual and predicted gaze positions in the XY-plane.

\subsection{5   Results}

As shown in Table \ref{table:model_accuracies}, the previous highest achievement on the EEGEyeNet dataset's absolute position task was an RMSE (Root Mean Square Error) of $55.4\pm 0.2$ mm, as reported by \cite{midterm-eeg-vit}. The results from all three methods, as described in Table \ref{table:model_accuracies}, demonstrated improved performance in terms of RMSE. In Method 1, where we implemented depthwise separable convolution, we achieved 53.5 mm. In Method 2, which applied 'EEGViT Trained with Clustered Data,' we achieved an RMSE of 53.4 mm. This result indicates the positive impact of data clustering on model accuracy. Finally, in Method 3, where we combined both methods by training the model with depthwise separable convolution on the clustered data, we achieved an even better RMSE of $51.6 \pm 0.2$ mm, reinforcing the effectiveness of these combined strategies.

\begin{figure}[h]
    \centering
    \includegraphics[width=.75\linewidth]{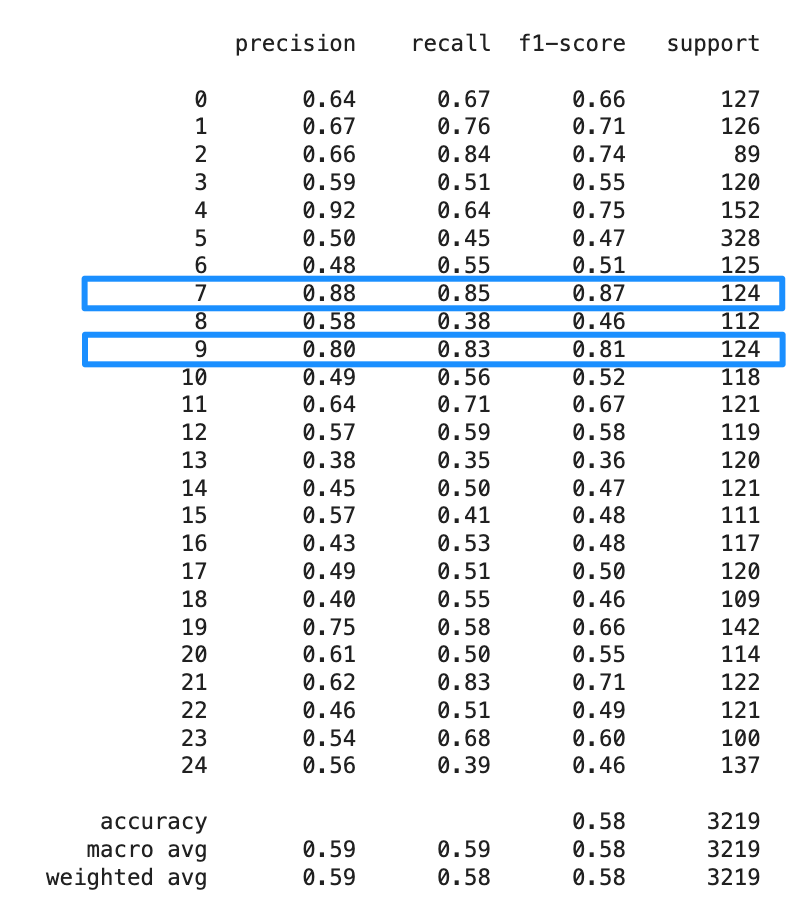}
    \caption{ \textbf{Classification Performance Metrics by Cluster:} This figure presents a detailed breakdown of classification metrics including precision, recall, F1-score, and support for 25 clusters, highlighting the performance of each cluster in the model evaluation.}
    \label{fig:Accuracies across clusters}
\end{figure}

\begin{figure}
    \centering
    \includegraphics[width=.9\linewidth]{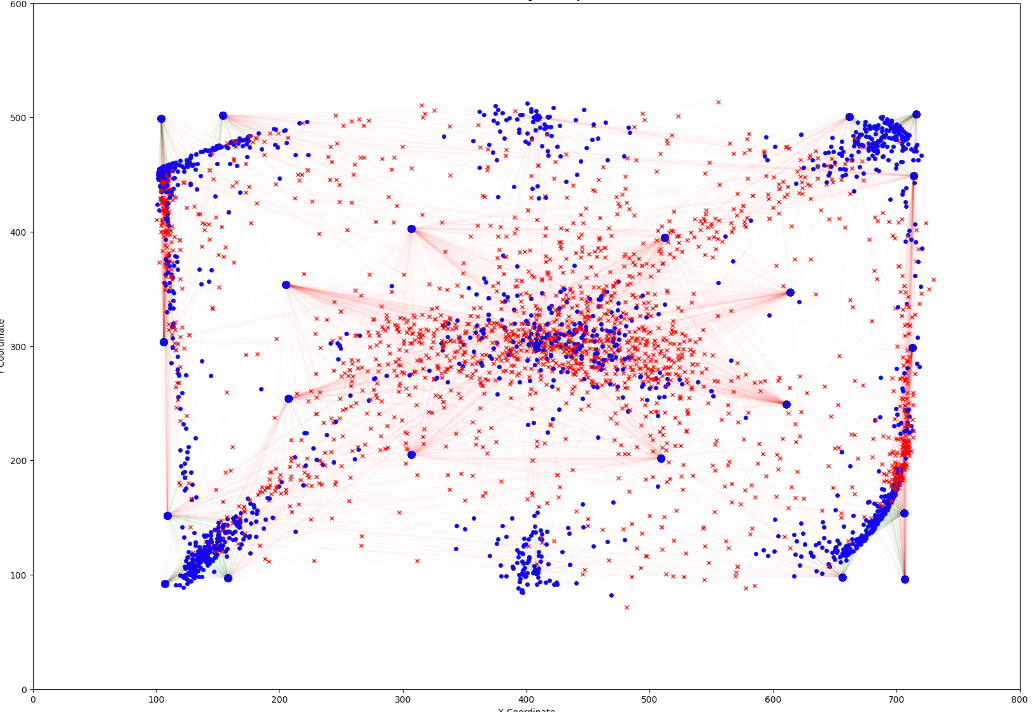}
    \caption{Visual of Test Error for Absolute Eye Position Showing Positions within 55.4 mm RMSE (Blue) and Positions Above 55.4 mm RMSE (Red). }
    \label{fig:visual-of-test-error}
\end{figure}

\begin{table}[h]
\centering
\begin{tabularx}{0.75\textwidth}{|X|c|} 
 \hline
 \textbf{Model} & \textbf{Absolute Position RMSE (mm)} \\ 
 \hline
 Naive Guessing & \makecell{123.3 $\pm$ 0.0} \\ 
 \hline
 CNN & \makecell{70.4 $\pm$ 1.1} \\
 \hline
 PyramidalCNN & \makecell{73.9 $\pm$ 1.9} \\
 \hline
 EEGNet & \makecell{81.3 $\pm$ 1.0} \\
 \hline
 InceptionTime & \makecell{70.7 $\pm$ 0.8} \\
 \hline
 Xception & \makecell{78.7 $\pm$ 1.6} \\
 \hline
 ViT - Base & \makecell{61.5 $\pm$ 0.6} \\
 \hline
 ViT - Base Pre-trained & \makecell{58.1 $\pm$ 0.6} \\
 \hline
 EEGViT & \makecell{61.7 $\pm$ 0.6} \\
 \hline
 EEGViT Pre - trained & \makecell{55.4 $\pm$ 0.2} \\
 \hline
 \hline
 Method 1 & \makecell{53.6 $\pm$ 0.6} \\ 
 \hline
 Method 2 & \makecell{53.4 $\pm$ 0.8} \\ 
 \hline
 \textbf{Method 3 (EEG-DCViT) } & \makecell{\textbf{51.6 $\pm$ 0.2}} \\
 \hline
\end{tabularx}

\caption{
\textbf{RMSE Comparisons for Absolute Position Task:} Root Mean Squared Error (RMSE) was converted to millimeters at a ratio of 2 pixels/mm. Lower RMSE values signify better accuracy, aligning closer to true values. Displayed values represent the average and standard deviation from 5 trials. \cite{midterm-eeg-vit}.}
\label{table:model_accuracies}
\end{table}
 
\begin{figure}[ht]
    \centering
    \includegraphics[width=.9\linewidth]{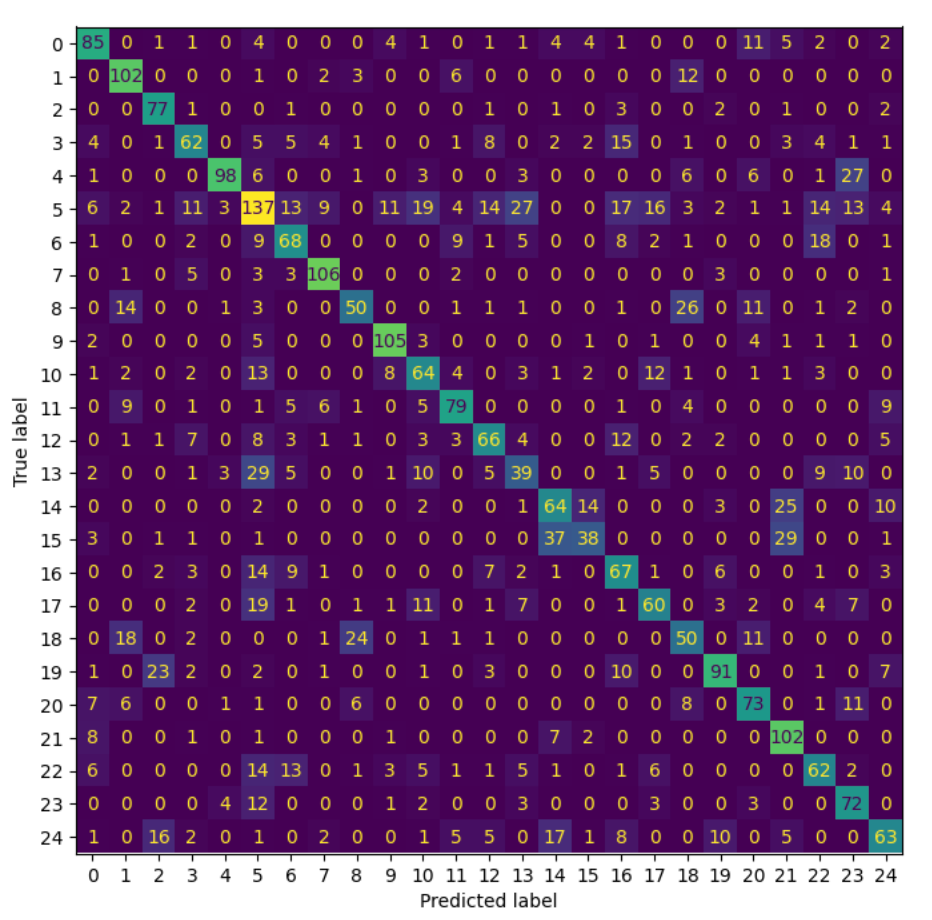}
    \caption{ \textbf{Confusion matrix across 25 clusters:} On the x-axis, we have the predicted values, which represent the outcomes as forecasted by our model. The y-axis, on the other hand, displays the true labels for each data point.
}
    \label{fig:heatmap}
\end{figure}

\subsection{6   Discussion}

These results collectively suggest that specialized training involving data clustering and DS-CNNs can significantly improve the accuracy of deep learning models in estimating absolute positions from EEG data.

\subsubsection{Computational Complexity:}
Traditionally, adding depth to vision transformers by increasing the number of convolutional layers adds computational complexity. Although our work does not include a comprehensive analysis, EEGViT has 86.0M trainable parameters, while EEG-DCViT has 86.2M trainable parameters. This results in insignificant differences in training time and memory usage.

The clustering technique runs in $\mathcal{O}(ndki)$ where n is the number of points, k is the number of clusters, d is the dimensionality of x, and i is the number of iterations that the algorithm takes to converge. In this case, the number of clusters is 25 and the number of dimensions is 2. Since these are constant, our algorithm runs in $\mathcal{O}(ni)$. Given only 21,000 data points and the hardware requirements to train EEG-DCViT, the algorithm converges within seconds.

\subsubsection{Understanding Test Error:}

One of the pivotal aspects of our study was the introduction of new visualization techniques that will help both computer scientists and neuroscientists understand the test error. During our training, we discovered a way to better understand the test error. Where is the test error coming from? Which eye positions have more error? We created a new visual in order to help us answer these questions (See Figure \ref{fig:visual-of-test-error}). For example, in Figure \ref{fig:visual-of-test-error}, we see that the eye positions on the top left and bottom right are more difficult for the model to perform well on compared to the bottom left and upper right-hand corners. Insights from neuroscientists and other subject matter experts will be critical in order to improve performance in these positions. In this same figure, faint lines between test locations and true labels show the distance between the target and predicted values. Notably, there are fewer red lines between the "inner" positions and the "outer" positions. This could mean that the model is good at determining the difference between someone looking at the center of the screen as opposed to the outside of the screen, though we did not quantify these results.

\subsubsection{Understanding EEG-ViT Performance:}
In order to understand the original EEG-ViT model, our team expanded the use of clustered eye positions shown in Figure \ref{fig:The centroids} by converting the model into a classifier. So, instead of predicting a location on a screen, the adjusted classification model would predict one of the 25 centroids shown in Figure \ref{fig:The centroids}. 

In the given classification report in figure \ref{fig:Accuracies across clusters}, the original EEGViT model's discriminative ability is quantified across multiple classes, with individual performance metrics presented for each class. Precision, recall, and F1-scores are provided, alongside the 'support' column, which denotes the actual number of samples for each respective class. Classes 7 (participant looking straight down) and 9 (participant looking straight up) are noteworthy, with F1-scores of 0.87 and 0.81 respectively, indicating a robust predictive performance for these categories. However, there are classes with notably lower F1-scores, such as class 13, indicating potential areas for model improvement. Similarly, 
the confusion matrix in Figure \ref{fig:heatmap} reveals that categories 7 and 9 closely match their predictions with the true labels, while class 13 has the least number of matched predictions. The high number of matched predictions in category 5 is attributed to its larger sample size in the dataset. Notably, the central category, represented three times more frequently than others, may skew the model's predictive distribution. Future iterations of the model could benefit from a more targeted approach in feature engineering and class-specific parameter tuning to uplift the predictive accuracy for underperforming classes.

\begin{figure}
    \centering
    \includegraphics[width=1\linewidth]{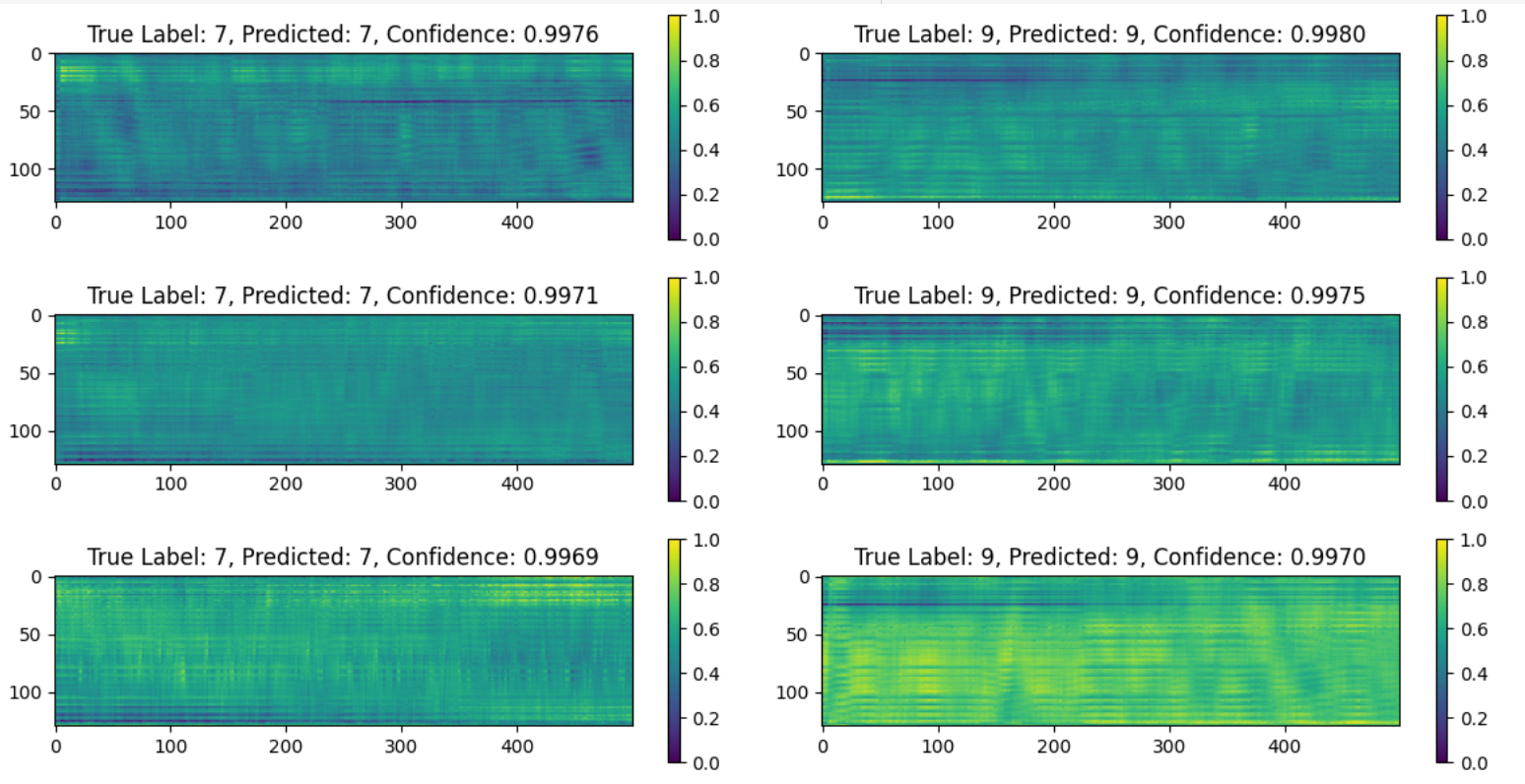}
    \caption{\textbf{Heat Map of EEG Data Samples Predicted with High Confidence by EEG-ViT Model:} This figure shows samples with high confidence from class 7 (left) and class 9 (right). Class 7 displays similar bright spot in the upper left corner. Class 9 displays a similar, larger, dark spot in the upper left.}
    \label{fig:eeg-heatmap}
\end{figure}

Furthermore, our team evaluated samples that were predicted with high confidence scores by EEGViT. We hypothesized that this would uncover patterns detected by the EEGViT model that are also interpretable to the human eye. We discovered similarities in samples classified with high confidence using EEG-ViT. As shown in Figure \ref{fig:eeg-heatmap}, there are clear similarities in the EEG samples. The cause of these similarities is undetermined; it could be due to leakage from ocular artifacts or valuable data that requires further insight from neuroscientists.

These visual tools not only facilitated a deeper understanding of the model's performance but also provided insights into the complex interplay of data features and model predictions. Visuals like this could be useful also as a communication tool between computer and neuroscientists. However, there is still a vast scope for innovation in this domain. Future research can focus on developing more advanced visualization techniques and tools that can provide even deeper insights into the workings of EEG data analysis models. This direction holds the promise of not only enhancing the interpretability of complex models but also fostering a more collaborative and intuitive approach to understanding neuroscience data.

Other deep learning approaches,\cite{an2023transfer,an2023survey,lu2022cot,lu2023machine,qiu2023modal},  particularly those applied in clinical image recognition \cite{tang2023active,xu2023metagrad,zhao2024deep} could also be explored to enhance the predictive accuracy in this experiment.

Our investigation applied enhanced pre-processing strategies and architectural improvements to a pre-trained EEG-ViT model, resulting in notable performance enhancements. The integration of vision transformers with EEG data analysis in our EEG-ViT model has demonstrated a powerful synergy. Importantly, the potential of these pre-processing techniques, when applied to Convolutional Neural Networks (CNNs), should not be overlooked. Future studies could explore how these strategies might elevate the performance of CNNs in EEG data analysis without the addition of a vision transformer model. This approach could offer valuable comparative insights between Transformer-based and convolutional architectures. Another promising avenue for research involves the use of Generative Adversarial Networks (GANs) in generating synthetic EEG datasets. This could potentially address the challenges of data scarcity and diversity in EEG analysis.

\subsection{7   Conclusion }

In conclusion, the deployment of pre-processing and using DS-CNNs has improved the performance of EEG-based predictive models. Our proposed model, in particular, has established new state-of-the-art results, achieving a benchmark RMSE of 51.6 mm. We are optimistic that the significant performance leap made by our model will serve as a cornerstone for future developments in EEG-based brain-computer interfaces and machine learning, inspiring continued innovation and research in the field.

\begin{credits}
 \subsubsection{\ackname} This study was part of the authors' work in the course "CSCI 6907 Applied Machine Learning" in The George Washington University.

\subsubsection{\discintname}
The authors declare no competing interests. 
\end{credits}


\bibliography{citations}




\end{document}